\RequirePackage{fix-cm}
\documentclass[smallextended]{svjour3}       
\smartqed  
\usepackage[numbers,sort&compress]{natbib}
\usepackage{amsmath,amssymb}
\usepackage{graphicx}
\usepackage{mathrsfs}
\usepackage[colorlinks,linkcolor=blue, anchorcolor=blue, citecolor=blue]{hyperref}
\usepackage{soul}
\usepackage{multirow}
\begin{document}

\title{$ A^2 $-FPN for Semantic Segmentation of Fine-Resolution Remotely Sensed Images
}


\author{Rui Li         \and
        Shunyi Zheng         \and
        Ce Zhang         \and
        Chenxi Duan         \and
        Libo Wang 
}


\institute{R. Li \at
School of Remote Sensing and Information Engineering, Wuhan University, Wuhan 430079, China \\
              \email{lironui@whu.edu.cn} \and
           S. Zheng \at
School of Remote Sensing and Information Engineering, Wuhan University, Wuhan 430079, China \\
              \email{syzheng@whu.edu.cn} \and
           C. Zhang \at
Lancaster Environment Centre, Lancaster University, Lancaster LA1 4YQ, United Kingdom; UK Centre for Ecology and Hydrology, Library Avenue, Lancaster, LA1 4AP, United Kingdom \\
              \email{c.zhang9@lancaster.ac.uk}  \and
           C. Duan \at
State Key Laboratory of Information Engineering in Surveying, Mapping, and Remote Sensing, Wuhan University, 430079, China \\
              \email{chenxiduan@whu.edu.cn}    \and
           L. Wang \at
School of Remote Sensing and Information Engineering, Wuhan University, Wuhan 430079, China \\
              \email{wanglibo@whu.edu.cn}  
}

\date{Received: date / Accepted: date}

\maketitle

\begin{abstract}
Semantic segmentation using fine-resolution remotely sensed images plays a critical role in many practical applications, such as urban planning, environmental protection, natural and anthropogenic landscape monitoring, etc. However, the automation of semantic segmentation, i.e., automatic categorization/labeling and segmentation is still a challenging task, particularly for fine-resolution images with huge spatial and spectral complexity. Addressing such a problem represents an exciting research field, which paves the way for scene-level landscape pattern analysis and decision making. In this paper, we propose
an approach for automatic land segmentation based on the Feature Pyramid Network (FPN). As a classic architecture, FPN can build a feature pyramid with high-level semantics throughout. However, intrinsic defects in feature extraction and fusion hinder FPN from further aggregating more discriminative features. Hence, we propose an Attention Aggregation Module (AAM) to enhance multi-scale feature learning through attention-guided feature aggregation. Based on FPN and AAM, a novel framework named Attention Aggregation Feature Pyramid Network ($ A^2 $-FPN) is developed for semantic segmentation of fine-resolution remotely sensed images. Extensive experiments conducted on three datasets demonstrate the effectiveness of our $ A^2 $-FPN in segmentation accuracy. Code is available at \url {https://github.com/lironui/A2-FPN}.
\end{abstract}

\section{Introduction}
\label{intro}
Semantic segmentation of remotely sensed images (i.e., assigns each pixel in images with a particular land cover category) has become one of the most important techniques for ground feature interpretation \cite{li2021abcnet}. The application of remotely sensed semantic segmentation is diverse in applied domains, such as urban planning, environmental protection, natural and anthropogenic landscape monitoring, etc. \cite{zhang2019joint, tong2020land, zhu2017deep}. Further, land cover information can provide insights from a panoramic perspective to help address socioeconomic and environmental grand challenges, such as food insecurity, poverty, climate change, and disaster risks. With recent advances in Earth observation technology, a constellation of satellite and airborne platforms have been launched \cite{duan2020thick, zhang2020scale}, by which substantial fine-resolution remotely sensed images are available for semantic segmentation. Traditional vegetation indices extracted from multi-spectral/multi-temporal images are used frequently as features to manifest the physical properties of land cover. However, the adaptability and flexibility of these indices are severely limited by their dependency upon hand-crafted feature descriptors \cite{li2020classification, gu2020semi}.

\par With rapid development in deep learning and deep convolutional neural networks (CNNs) in particular, significant breakthroughs of semantic segmentation in remote sensing have been witnessed in recent years \cite{su2021improved, wang2021novel, wang2021transformer}. Compared with vegetation indices, CNNs can harness a wide variety of information, such as spectral characteristics, spatial context, and the interaction among different land cover categories. Most importantly, CNNs have the strong ability to learn nonlinear and hierarchical features automatically, which forms the end-to-end framework from the raw image to meaningful information and insights directly \cite{tong2021sat, wang2021novel, zhang2021lightweight}.

\par The scale variation of geospatial objects is general in remote sensing imagery, particularly with the fine-resolution. Therefore, the multi-scale representation is of great importance for dealing with such problem. Feature Pyramid Network (FPN) \cite{lin2017feature} is a widely-used framework to address the issue of multi-scale processing. By fusing adjacent features through lateral connections and the top-down pathway, FPN constructs a feature pyramid with strong semantics
at all scales, thereby leveraging the inherent feature hierarchy.

\par Although effective in multi-scale feature representations, the designs of FPN hinder feature pyramids from further aggregating more discriminative features for segmentation. Specifically, in the procedure of feature fusion, feature maps are upsampled and fused directly, losing the rich context information. To remedy the defect of FPN, we propose an Attention Aggregation Module (AAM) based on the linear attention mechanism \cite{li2021multistage} to enahnce multi-scale feature learning through attention-guided feature aggregation, thereby designing $ A^2 $-FPN. Compared to mainstream encoder-decoder frameworks, $ A^2 $-FPN is distinctive in two significant aspects: (1) It encodes semantic features form multi-scale layers; (2) It extracts discriminative features by extracting global context information. 

\section{Related Work}
\label{sec:rew}
\subsection{Feature Pyramid Network}
\begin{figure}[ht]
\centering
\includegraphics[scale=0.2]{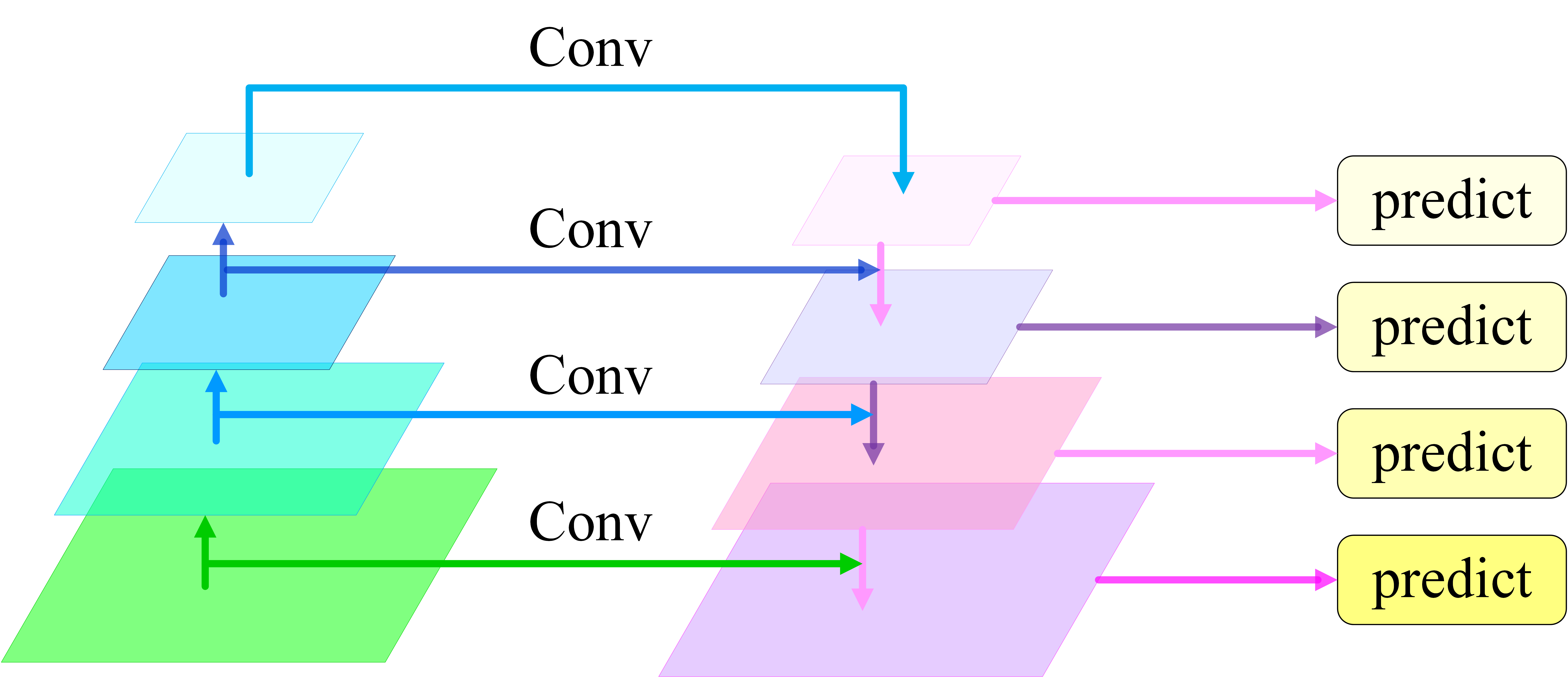}
\caption{Illustration of the architecture of Feature Pyramid Network for detection.}
\label{fig:1}
\end{figure}

\noindent The feature pyramid network (FPN) is initially designed for object detection, aiming at leveraging the pyramidal feature hierarchy \cite{lin2017feature}. The components of the FPN are comprised of a bottom-up pathway, a top-down pathway, and lateral connections, as illustrated in Fig. \ref{fig:1}. The bottom-up pathway takes the ResNet as the backbone \cite{he2016deep}, where the feature hierarchy is computed with feature maps being generated at multiple scales. The feature maps at top pyramid levels are spatially coarse but with high-level semantics. The top-down pathway interpolates fine-resolution features by upsampling from high-level feature maps, which are merged and refined with features at the same spatial size from the bottom-up pathway via lateral connections. The effectiveness of FPN has been demonstrated in several applications, including object detection \cite{lin2017feature}, panoptic segmentation \cite{kirillov2019panoptic}, and super-resolution \cite{shoeiby2020mosaic}.

\subsection{Semantic Segmentation}
\label{sec:ss}
The goal of semantic segmentation from fine-resolution remotely sensed images is to assign each pixel of imagery an accurate ground object class. After the first successful Fully Convolutional Network (FCN), deep learning methods have been successfully and extensively introduced and applied to the remotely sensed images especially for those with the fine-resolution. Meanwhile, semantic segmentation has shown great potential for practical applications in remote sensing areas including road detection \cite{wei2020simultaneous, shamsolmoali2020road}, urban resource management \cite{zhang2020identifying, li2020crop}, and land-use mapping \cite{tu2020regional}. For example, a novel CNN-based multi-stage framework is introduced by \cite{wei2020simultaneous} to extract road surface and centerline tracing simultaneously. \cite{zhang2020identifying} characterizes and classifies individual plants based on semantic segmentation methods by continuously increasing patch scale. The recently developed semantic segmentation approaches using deep learning create a new paradigm for land-use mapping \cite{tu2020regional}.

\subsection{The Attention Mechanism}
\label{sec:am}
\begin{figure*}[ht]
\centering
\includegraphics[scale=0.5]{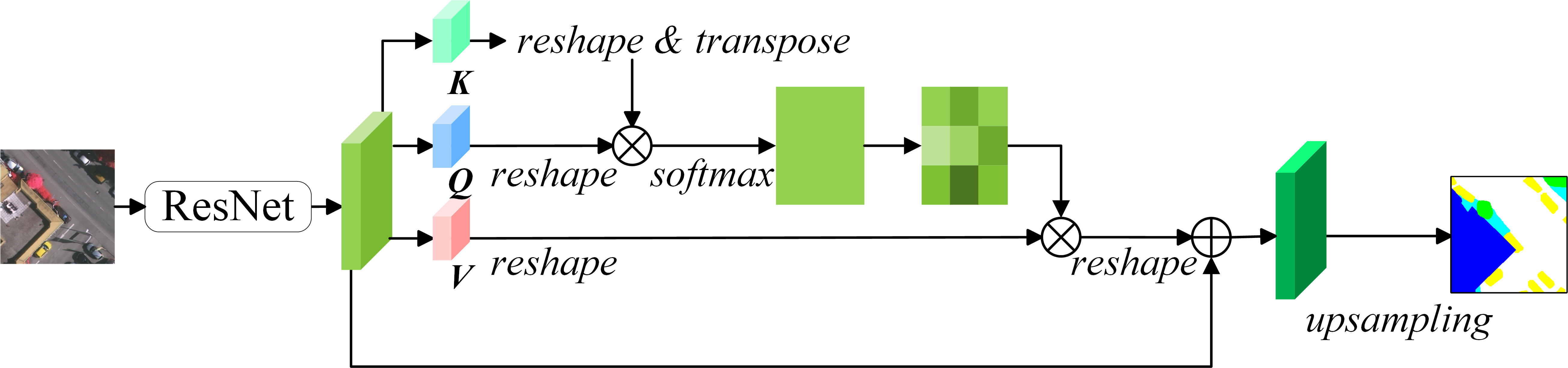}
\caption{Illustration of the architecture of dot-product attention mechanism.}
\label{fig:2}
\end{figure*}

\noindent The accuracy of segmentation relies on inference from sufficient context information. To this end, the dot-product attention mechanism is introduced to capture the global context information. However, the high memory and computational consumptions increase quadratically with the input size, which heavily hinders the actual application of the dot-product attention mechanism. Here, we illustrate the principles of the dot-product attention mechanism as well as the attempts to reduce the complexity of the attention mechanism, especially the linear attention mechanism utilized in the proposed $ A^2 $-FPN. By default, vectors in this section refer to column vectors.
\subsubsection{The Dot-Product Attention Mechanism}
The height, weight, and channels of the input are denoted as  $ H $, $ W $ and $ C $, respectively. $ \boldsymbol {X}=[\boldsymbol {x}_1, \boldsymbol {x}_2,...,\boldsymbol {x}_N] \in \mathbb{R}^{N \times C} $ refers to the input feature, where $ N=H \times W $. First, the dot-product attention mechanism uses three projected matrices
$ \boldsymbol{W}_q \in \mathbb{R}^{D_x \times D_k} $, $ \boldsymbol{W}_k \in \mathbb{R}^{D_x \times D_k} $, and $ \boldsymbol{W}_v \in \mathbb{R}^{D_x \times D_v} $ to obtain the \emph{query} matrix $ \boldsymbol Q $, \emph{key} matrix $ \boldsymbol K $  and \emph{value} matrix $ \boldsymbol V $ as:
\begin{equation} \label{equa:1}
\begin{split}
\boldsymbol {Q = XW_q} \in \mathbb{R}^{N \times D_k},\\
\boldsymbol {K = XW_k} \in \mathbb{R}^{N \times D_k},\\
\boldsymbol {V = XW_v} \in \mathbb{R}^{N \times D_v}.
\end{split}
\end{equation} 
$ \boldsymbol Q $ and $ \boldsymbol K $ are identical in their shapes. To compute the similarity between the $ i $-th \emph{query} feature $ {\boldsymbol q}_i^T \in \mathbb{R}^ {D_k} $ and the $ j $-th \emph{key} feature $ {\boldsymbol k}_j \in \mathbb{R}^{D_k} $, a normalization function $ \rho $ is adopted as $ \rho({\boldsymbol q}_i^T \cdot {\boldsymbol k}_j) \in \mathbb{R}^1 $. Thereafter, similarities between all pairs of pixels are computed and taken as weights. The output is generated by aggregating all positions using weighted summation:

\begin{equation} 
D(\boldsymbol {Q, K, V}) = \rho (\boldsymbol Q \boldsymbol {K}^T) \boldsymbol V. \label{equa:2}
\end{equation}

\par The normalization function is set as softmax:
\begin{equation}
\rho (\boldsymbol Q \boldsymbol {K}^T) = {\rm softmax}_{row}(\boldsymbol Q \boldsymbol {K}^T). \label{equa:3}
\end{equation}
where $ {\rm softmax}_{row} $ denotes that the softmax is operated along the row of matrix $ \boldsymbol Q \boldsymbol {K}^T  $. The global context information is captured by the $\rho (\boldsymbol Q \boldsymbol {K}^T) $
through the modeling of the similarities among all pairs of pixels in the input. However, as $ \boldsymbol {Q} \in \mathbb{R}^{N \times D_k} $ and $ \boldsymbol {K}^T \in \mathbb{R}^{D_k \times N} $, the multiplication between $ \boldsymbol Q $ and $ \boldsymbol K^T $ belongs to $ \mathbb{R}^{N \times N} $, leading to the $ O(N^2) $ time and memory complexity.
 
\par The dot-product attention mechanism is initially proposed for machine translation \cite{vaswani2017attention}, which is introduced and modified for computer vision by the non-local module (Fig. \ref{fig:2}). Based on the dot-product attention mechanism and its variants, attention-based networks have been proposed to tackle the semantic segmentation task, such as Dual Attention Network (DANet) \cite{fu2019dual}, Object Context Network (OCNet) \cite{yuan2018ocnet}, and Co-occurrent Feature Network (CFNet) \cite{zhang2019co}.

\subsubsection{Generalization and Simplification}
\par Given the normalization function is softmax, the $ i $-th row in the output matrix produced by the dot-product attention mechanism can be written as:
\begin{equation}
D(\boldsymbol {Q, K, V})_i = \frac{\sum_{j=1}^N e^{\boldsymbol {q}_i^T \cdot \boldsymbol {k}_j} \boldsymbol{v}_j}{\sum_{j=1}^N e^{\boldsymbol{q}_i^T \cdot \boldsymbol{k}_j}}. \label{equa:4}
\end{equation}

\par Equation \ref{equa:4} can be generalized into any normalization function as:
\begin{equation}
D(\boldsymbol {Q, K, V})_i = \frac{\sum_{j=1}^N {\rm sim}({\boldsymbol {q}_i, \boldsymbol {k}_j}) \boldsymbol{v}_j}{\sum_{j=1}^N {\rm sim} ({\boldsymbol{q}_i, \boldsymbol{k}_j})},
{\rm sim}({\boldsymbol {q}_i, \boldsymbol {k}_j}) \ge 0, \label{equa:5}
\end{equation}
$ {\rm sim}({\boldsymbol {q}_i, \boldsymbol {k}_j}) $ depicts the similarity between the $ \boldsymbol {q}_i $ and $ \boldsymbol {k}_j $, which can be expanded as $ {\rm sim}({\boldsymbol {q}_i, \boldsymbol {k}_j}) = \phi (\boldsymbol {q}_i)^T \varphi (\boldsymbol {k}_j) $. We can further rewrite equation \ref{equa:4} to equation \ref{equa:6} and then simplify it as equation \ref{equa:7}:
\begin{equation}
D(\boldsymbol {Q, K, V})_i = \frac{\sum_{j=1}^N \phi({\boldsymbol {q}_i)^T \varphi(\boldsymbol {k}_j}) \boldsymbol{v}_j}{\sum_{j=1}^N \phi({\boldsymbol {q}_i)^T \varphi(\boldsymbol {k}_j})}, \label{equa:6}
\end{equation}

\begin{equation}
D(\boldsymbol {Q, K, V})_i = \frac{\phi({\boldsymbol {q}_i)^T\sum_{j=1}^N  \varphi(\boldsymbol {k}_j}) \boldsymbol{v}_j}{\phi({\boldsymbol {q}_i)^T\sum_{j=1}^N  \varphi(\boldsymbol {k}_j})}. \label{equa:7}
\end{equation}

\par In particular, equation \ref{equa:5} is identical to equation \ref{equa:4}, when $ {\rm sim}({\boldsymbol {q}_i, \boldsymbol {k}_j}) = e^{\boldsymbol{q}_i^T \cdot \boldsymbol{k}_j} $. The equation \ref{equa:7} can be represented as the vectorized form:

\begin{equation}
D(\boldsymbol {Q, K, V}) = \frac{{\rm \phi}(\boldsymbol Q){\rm \varphi} (\boldsymbol K)^T \boldsymbol{V}}{{\rm \phi}(\boldsymbol {Q}) \sum_j{{\rm \varphi} (\boldsymbol {K})_{i, j}^T }}, \label{equa:8}
\end{equation}

As $ {\rm sim}({\boldsymbol {q}_i, \boldsymbol {k}_j}) = \phi (\boldsymbol {q}_i)^T \varphi (\boldsymbol {k}_j) $ replaces the softmax function, the order of the commutative operation can be altered, thereby reducing the computationally intensive operations. Specifically, we can compute the multiplication between $ {\rm \varphi} (\boldsymbol K)^T $ and $ \boldsymbol{V} $ first and then multiply the result and $ {\rm \phi}  ({\boldsymbol Q}) $, resulting in only $ O(dN) $ time and memory complexity. The appropriate $ \rm \phi(\cdot) $ and $ \varphi (\cdot) $ and enable the drastically reduced computation without sacrificing the accuracy \cite{li2020multi, katharopoulos2020transformers}.
\subsubsection{The Linear Attention Mechanism}
\begin{figure}[ht]
\centering
\includegraphics[scale=0.6]{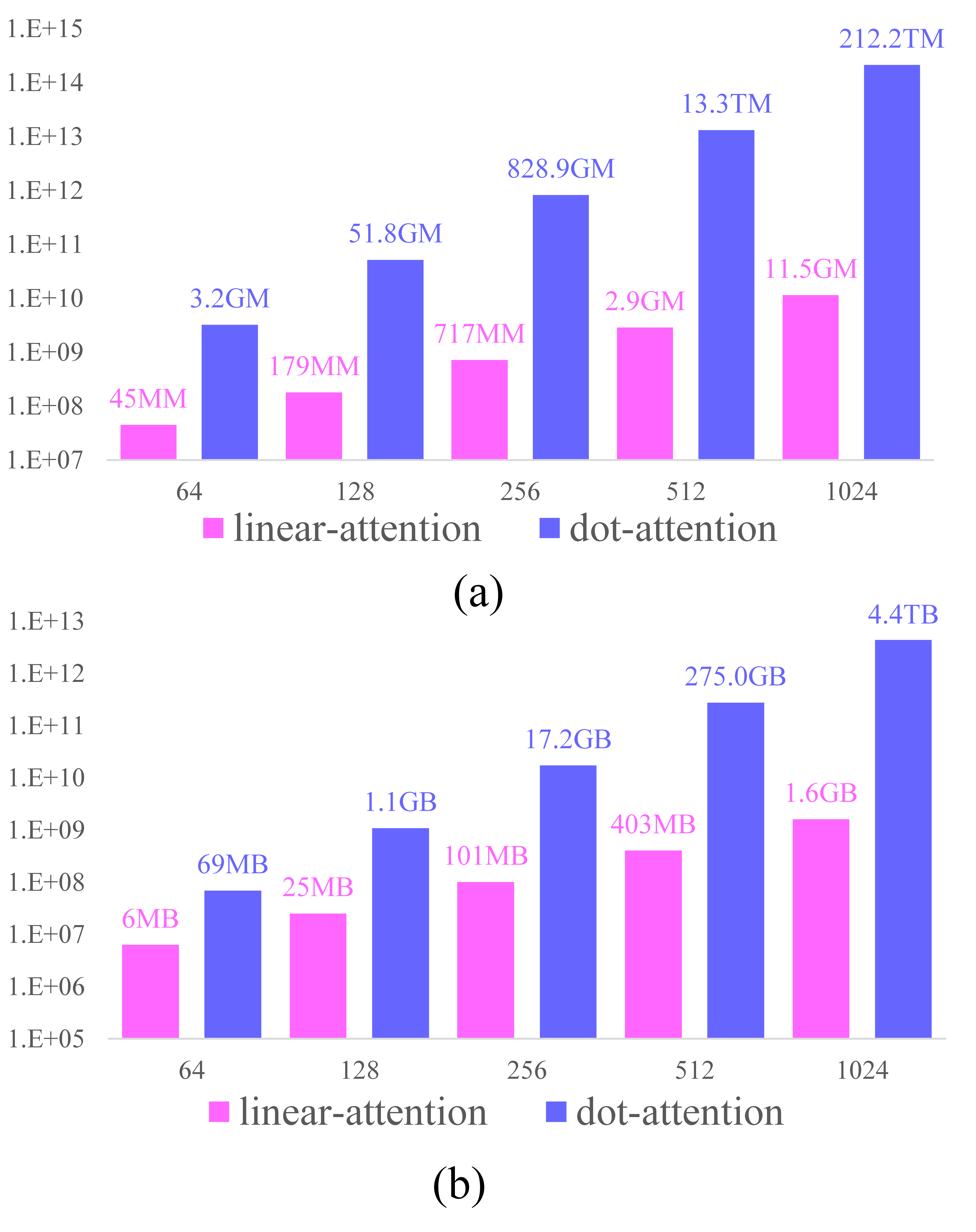}
\caption{The (a) computation requirement and (b) memory requirement between the linear attention mechanism and dot-product attention mechanism under different input sizes. The calculation assumes. The calculation assumes $ D = D_v = 2D_k = 64 $. MM denotes 1 Mega multiply-accumulation (MACC), where 1 MACC means 1 multiplication and 1 addition operation. GM means 1 Giga MACC, while TM signifies 1 Tera MACC. Similarly, MB, GB, and TB represent 1 MegaByte, 1 GigaByte, and 1 TeraByte, respectively. Note the figure is shown on the log scale.}
\label{fig:3}
\end{figure}

By replacing the softmax into its first-order approximation of Taylor expansion, we developed a linear attention mechanism in our previous research \cite{li2021multistage} as:
\begin{equation}
e^{\boldsymbol {q}_i^T \cdot \boldsymbol {k}_j} \approx 1 + \boldsymbol {q}_i^T \cdot \boldsymbol {k}_j, \label{equa:9}
\end{equation}

\par However, the above approximation cannot guarantee the non-negative property of the normalization function. Hence, we normalize $ \boldsymbol {q}_i $ and $ \boldsymbol {k}_j $ by $ l_2 $ norm to ensure $ \boldsymbol {q}_i^T \cdot \boldsymbol {k}_j \geq -1 $:
\begin{equation}
{\rm sim}({\boldsymbol {q}_i, \boldsymbol {k}_j}) = 1 + (\frac{\boldsymbol {q}_i}{\parallel \boldsymbol {q}_i \parallel}_2)^T (\frac{\boldsymbol {k}_j}{\parallel \boldsymbol {k}_j \parallel}_2), \label{equa:10}
\end{equation}

\par We then rewrite equation \ref{equa:5} into equation \ref{equa:11}, and simplify it into equation \ref{equa:12}:

\begin{equation}
D(\boldsymbol {Q, K, V})_i = \frac{\sum_{j=1}^N (1 + (\frac{\boldsymbol {q}_i}{\parallel \boldsymbol {q}_i \parallel}_2)^T (\frac{\boldsymbol {k}_j}{\parallel \boldsymbol {k}_j \parallel}_2)) \boldsymbol{v}_j}{\sum_{j=1}^N (1 + (\frac{\boldsymbol {q}_i}{\parallel \boldsymbol {q}_i \parallel}_2)^T (\frac{\boldsymbol {k}_j}{\parallel \boldsymbol {k}_j \parallel}_2))}, \label{equa:11}
\end{equation}

\begin{equation}
D(\boldsymbol {Q, K, V})_i = \frac{\sum_{j=1}^N \boldsymbol{v}_j + (\frac{\boldsymbol {q}_i}{\parallel \boldsymbol {q}_i \parallel}_2)^T \sum_{j=1}^N (\frac{\boldsymbol {k}_j}{\parallel \boldsymbol {k}_j \parallel}_2) \boldsymbol{v}_j^T}{N + (\frac{\boldsymbol {q}_i}{\parallel \boldsymbol {q}_i \parallel}_2)^T \sum_{j=1}^N (\frac{\boldsymbol {k}_j}{\parallel \boldsymbol {k}_j \parallel}_2)}. \label{equa:12}
\end{equation}
The vectorized form of equation \ref{equa:12} is:
\begin{equation}
D(\boldsymbol {Q, K, V}) = \frac{\sum_{j} \boldsymbol{V}_{i,j} + (\frac{\boldsymbol {Q}}{\parallel \boldsymbol {Q} \parallel}_2) ((\frac{\boldsymbol {K}}{\parallel \boldsymbol {K} \parallel}_2)^T \boldsymbol{V})}{N + (\frac{\boldsymbol {Q}}{\parallel \boldsymbol {Q} \parallel}_2) \sum_{j} (\frac{\boldsymbol {K}}{\parallel \boldsymbol {K} \parallel}_2)_{i,j}^T}. \label{equa:13}
\end{equation}

As $ \sum_{j=1}^N (\frac{\boldsymbol {k}_j}{\parallel \boldsymbol {k}_j \parallel}_2) \boldsymbol{v}_j^T $ and $ \sum_{j=1}^N (\frac{\boldsymbol {k}_j}{\parallel \boldsymbol {k}_j \parallel}_2) $ could be computed only once and reused for each query, time and space complexity of the linear attention mechanism based on equation \ref{equa:13} is $ O(dN) $. Specifically, given a feature $ \boldsymbol {X}=[\boldsymbol {x}_1, \boldsymbol {x}_2,...,\boldsymbol {x}_N] \in \mathbb{R}^{N \times C} $, both the dot-attention and linear attention generate the \emph{query} matrix $ \boldsymbol Q $, \emph{key} matrix $ \boldsymbol K $  and \emph{value} matrix $ \boldsymbol V $. For the dot-attention, the $ N \times N $ matrix is generated by multiplying the transposed \emph{key} matrix $ \boldsymbol K $ and the \emph{value} matrix $ \boldsymbol V $, resulting in $ O(D_k N^2) $ time complexity and $ O(N^2) $ space complexity to compute the similarity using the softmax function. Thus, the dot-attention would occupy at least $ O(N^2) $ memory and require $ O(D_k N^2) $ computation to calculate the similarity between each pair of positions. For linear attention, as the softmax function is substituted for the first-order approximation of Taylor expansion, we can alter the order of the commutative operation and avoid multiplication between the reshaped \emph{key} matrix $ \boldsymbol K $ and \emph{query} matrix $ \boldsymbol Q $. Therefore, we can calculate the product between $ \boldsymbol {K}^T $ and $ \boldsymbol V $ first and then multiply the result and $ \boldsymbol Q $ with only $ O(dN) $ time complexity and $ O(dN) $ space complexity. The concrete comparison can be seen in Fig. \ref{fig:3}.

\section{Attention Aggregation Feature Pyramid Network}
\label{sec:aafpn}
\begin{figure*}[ht]
\centering
\includegraphics[scale=0.35]{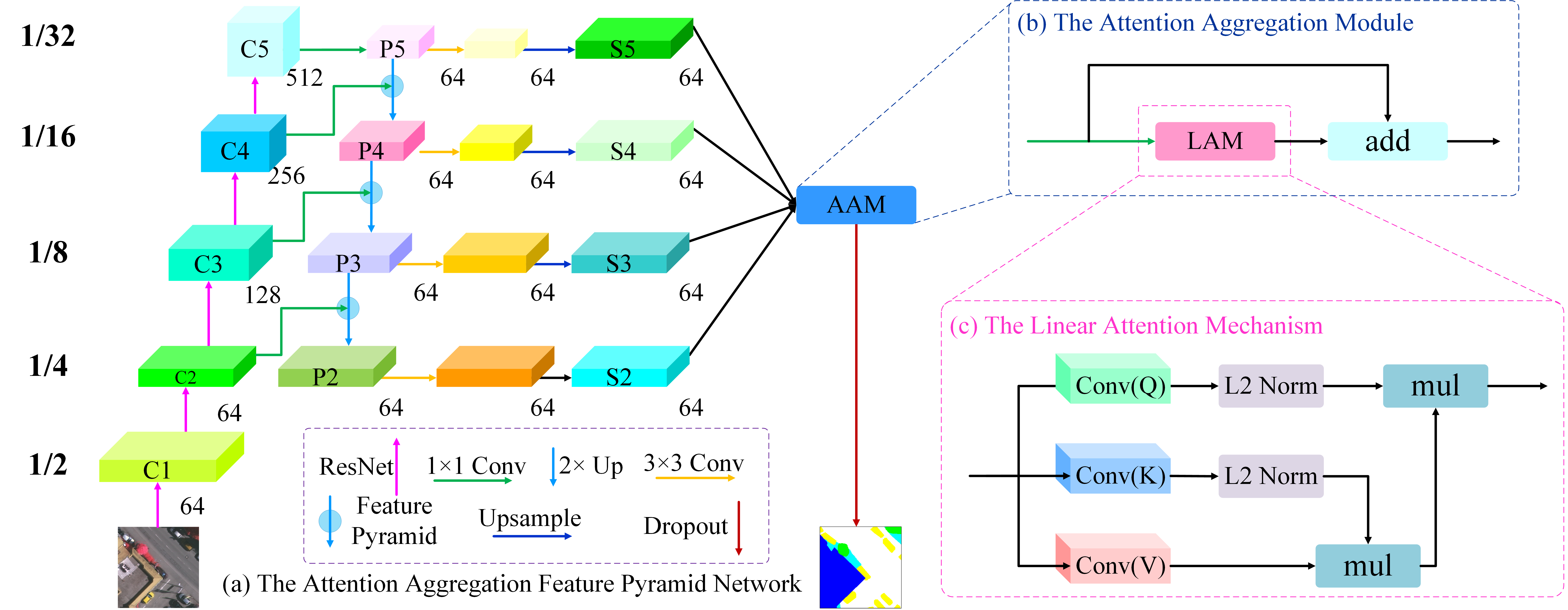}
\caption{The structure of (a) the overall framework of our $ A^2 $-FPN, (b) the Attention Aggregation Module, and (c) the Linear Attention Mechanism (taking the attention1 as an example). The figures (e.g., 64, 128, 512) near the features indicate the number of channels..}
\label{fig:4}
\end{figure*}

\noindent The overall framework of the proposed $ A^2 $-FPN is demonstrated in Fig. \ref{fig:4}. As a single end-to-end network, the major components of our $ A^2 $-FPN include the bottom-up pathway (i.e., the first column in Fig. \ref{fig:4}a), the top-down pathway (i.e., the second column in Fig.\ref{fig:4}a), the lateral connections (i.e., the $ 1 \times 1$ convolutional layer between the first and second column in Fig. \ref{fig:4}a), the feature pyramid (i.e., the second and third columns in Fig. \ref{fig:4}a), and the Attention Aggregation Module (i.e., Fig. \ref{fig:4}b). We will elaborate on each component below.

\subsection{The Bottom-up Pathway}
\label{sec:bup}
To design a simple and efficient framework, we select the ResNet-34 as the backbone of the bottom-up pathway rather than the complicated backbones such as ResNet-101. Based on ResNet-34, the bottom-up pathway conducts the feedforward learning and generates the feature hierarchy. The feature maps are generated at different spatial resolutions with a scaling step of 2. The top levels of feature maps have large spatial context with coarse resolution, whereas the bottom levels of feature maps present small context information with fine resolution. We use C2, C3, C4, and C5 to indicate the output feature map of each residual block in ResNets (see above Fig. \ref{fig:4}), while the spatial size of C2, C3, C4, and C5 are 1/4, 1/8, 1/16, and 1/32 of the input size, respectively. Due to its large memory footprint, C1 is not included in the pyramid.  

\subsection{The Top-Down Pathway and Lateral Connections}
\label{sec:tdp}
\begin{figure}[ht]
\centering
\includegraphics[scale=0.5]{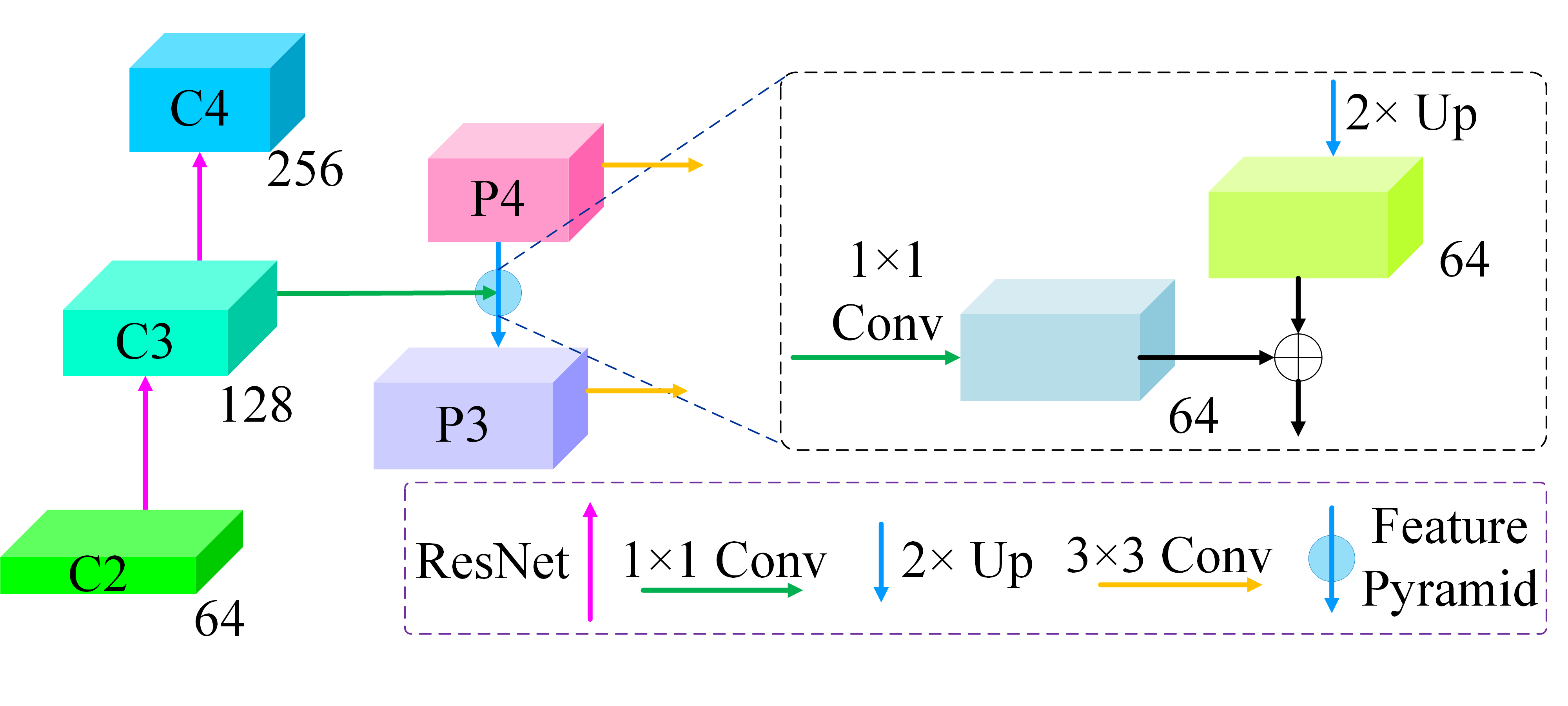}
\caption{The feature pyramid in the proposed $ A^2 $-FPN.}
\label{fig:5}
\end{figure}

\noindent The top-down pathway up-samples semantically rich but spatially coarse feature maps from top pyramid levels to create fine resolution features, which are then merged and refined with corresponding features from the bottom-up pathway via lateral connections. As shown in Fig. \ref{fig:5}, a top-down layer and a lateral connection constitute a feature pyramid in the proposed $ A^2 $-FPN. The generated feature maps are denoted as P2, P3, P4, and P5 accordingly. With a coarse resolution feature map (e.g., P4 in Fig. \ref{fig:5}), we up-sample its spatial resolution by a factor of 2 (the up-sampling mode is set as the nearest neighbor for simplicity). By element-wise addition, the up-sampled map is then fused with the corresponding map (with a $ 1 \times 1 $ convolutional layer to reduce dimensions of the channel) in the bottom-up pathway.

\par The above procedure is iterated until the finest resolution map is generated. To start the iteration, the coarsest resolution map (e.g., P5 in Fig. \ref{fig:4}) is directly produced by a $ 1 \times 1 $  convolutional layer on C5. After the merged map generated by the corresponding feature pyramid, a $ 3 \times 3 $  convolution is attached to produce the final feature map, where the aliasing effect caused by upsampling is mitigated. The feature pyramid combines low-level contextual information into spatial feature maps, which improves the representation capability of low-level side networks. Interpreting different scales of land covers requires different levels of context information. Indeed, a large spatial context is contained in the high-level features since the deep convolution layers have larger receptive fields than the shallow ones. Hence, when merged with high-level features, the low-level side networks acquire the multi-scale context information to improve its accuracy of segmentation.

\subsection{The Attention Aggregation Module}
\label{sec:aam}
The local-aware property severely limits the potential of the CNN to capture the global context information which is paramount for semantic segmentation of remotely sensed images. Graphical models and pyramid pooling modules partly remedy the context issue, however, the contextual dependencies for whole input regions are homogeneous and non-adaptive, ignoring the disparity between contextual dependencies and local representation of different categories. Besides, those strategies which are usually utilized only in one layer do not sufficiently leverage the long-range dependencies of feature maps.

\par FPN is an effective framework to address the multi-scale processing issue. However, the designs of FPN cause the lack of context information in feature maps. Here, to extract the global context information, we design the Attention Aggregation Module which enhances long-range dependencies of feature maps on multi-level based on the linear attention mechanism (LAM) (Fig. \ref{fig:4}b and Fig. \ref{fig:4}c). Specifically, the four feature maps (i.e., S2, S3, S4, and S5) generated by the corresponding feature pyramid are first concatenated and then fed into the $ 1 \times 1 $  convolutional layer. Thereafter, the linear attention mechanism is utilized to capture global context information and further refine fused feature maps. Finally, the refined features are added with the original concatenated features. 

\section{Experimental Results}
\label{sec:exper}

\subsection{Datasets}
We test the effectiveness of $ A^2 $-FPN based on the ISPRS Vaihingen and the Potsdam datasets (\url {http://www2.isprs.org/commissions/comm3/wg4/semantic-labeling.html}), as well as Gaofen Image Dataset (GID) \cite{tong2020land}. 

\par \textbf{Vaihingen:} 33 tiles are extracted from true orthophoto and the co-registered normalized digital surface models (nDSMs) in the Vaihingen dataset. The ground sampling distance (GSD) of tiles in Vaihingen is 9 cm and the average size is $ 2494 \times 2064 $ pixels. We utilize tiles: 2, 4, 6, 8, 10, 12, 14, 16, 20, 22, 24, 27, 29, 31, 33, 35, 38 for testing, tile: 30 for validation, and the remaining 15 images for training. We use the near-infrared, red, and green channels only without nDSM in our experiments.

\par \textbf{Potsdam:} The Potsdam dataset contains 38 tiles extracted from true orthophoto and the co-registered normalized DSMs. The GSD of tiles in Potsdam is 5 cm and the size of each tile is $ 6000 \times 6000 $. We utilize tiles: 2\_13, 2\_14, 3\_13, 3\_14, 4\_13, 4\_14, 4\_15, 5\_13, 5\_14, 5\_15, 6\_13, 6\_14, 6\_15, 7\_13 for testing, tile: 2\_10 for validation, and the remaining 22 tiles, except 7\_10 with error annotations, for training. The nDSM is not employed in our experiments.

\par \textbf{GID:} The GID contains 150 RGB images of $ 7200 \times 6800 $ pixels. Each image covers a geographic region of  $ 506 km^2 $ captured by the Gaofen 2 satellite sensor. Following the previous work \cite{li2021macu}, we select 15 images contained in GID, which cover the whole six categories. We partition each image into non-overlapping patch sets of size $ 512 \times 512 $ pixels. Thereafter, 50\% patches are selected randomly as the training set, 10\% patches are chosen as the validation set, and the remained 40\% patches are reserved as the test set.

\subsection{Evaluation Metrics}
\par The performance of our $ A^2 $-FPN, as well as comparative methods, is measured by the overall accuracy (OA), the mean Intersection over Union (mIoU), and the F1 score (F1). Based on the accumulated confusion matrix, the OA, mIoU, and F1 are computed as:
\begin{equation}
OA = \frac{\sum_{k=1}^NTP_k}{\sum_{k=1}^N TP_k+FP_k+TN_k+FN_k}, \label{equa:14}
\end{equation}

\begin{equation}
mIoU = \frac{1}{N} \sum_{k=1}^N \frac{TP_k}{TP_k+FP_k+FN_k}, \label{equa:15}
\end{equation}

\begin{equation}
F1 = 2 \times \frac{precision \times recall}{precision + recall}, \label{equa:16}
\end{equation}
where $ TP_k, FP_k, TN_k $ and $ FN_k $ indicate the true positive, false positive, true negative, and false negatives, respectively, for object indexed as class $ k $. OA is calculated for all categories including the background.

\subsection{Experimental Setting}
\par We implemented the proposed $ A^2 $-FPN and comparative algorithms using PyTorch under the Python platform and trained them using a single Tesla V100 with 32 batch size and Adam optimizer. The learning rate is parameterized as 0.0003. For training, we cropped the original tiles into $ 512 \times 512 $ patches and augmented them by rotating, resizing, horizontal axis flipping, vertical axis flipping, and adding random noise. The test time augmentation (TTA) in terms of rotating and flipping is applied for all algorithms accordingly.

\subsection{Results on the ISPRS Vaihingen Dataset}
We compare our method with seven existing methods on the Vaihingen test set and quantitative comparisons are shown in Table \ref{table:1}. For a fair comparison, the backbone of ResNet-based algorithms is set as ResNet-34 consistently. Our $ A^2 $-FPN outperforms other encoder-decoder methods (e.g., U-Net and CE-Net), attention-based methods (e.g., DANet), and context aggregation methods (e.g., PSPNet and EaNet) by a significant margin. To be specific, at least 1.6\% in mean F1 score, 0.6\% in OA, and 2.5\% in mIoU higher than the other comparative methods. Especially, the F1 score of Car predicted by our $ A^2 $-FPN is far higher than any other approaches, which increase the second-best CE-Net by a large margin of 5.7\%, demonstrating the effectiveness of the Attention Aggregation Module. 

\par
\begin{figure*}[ht]
\centering
\includegraphics[scale=1.3]{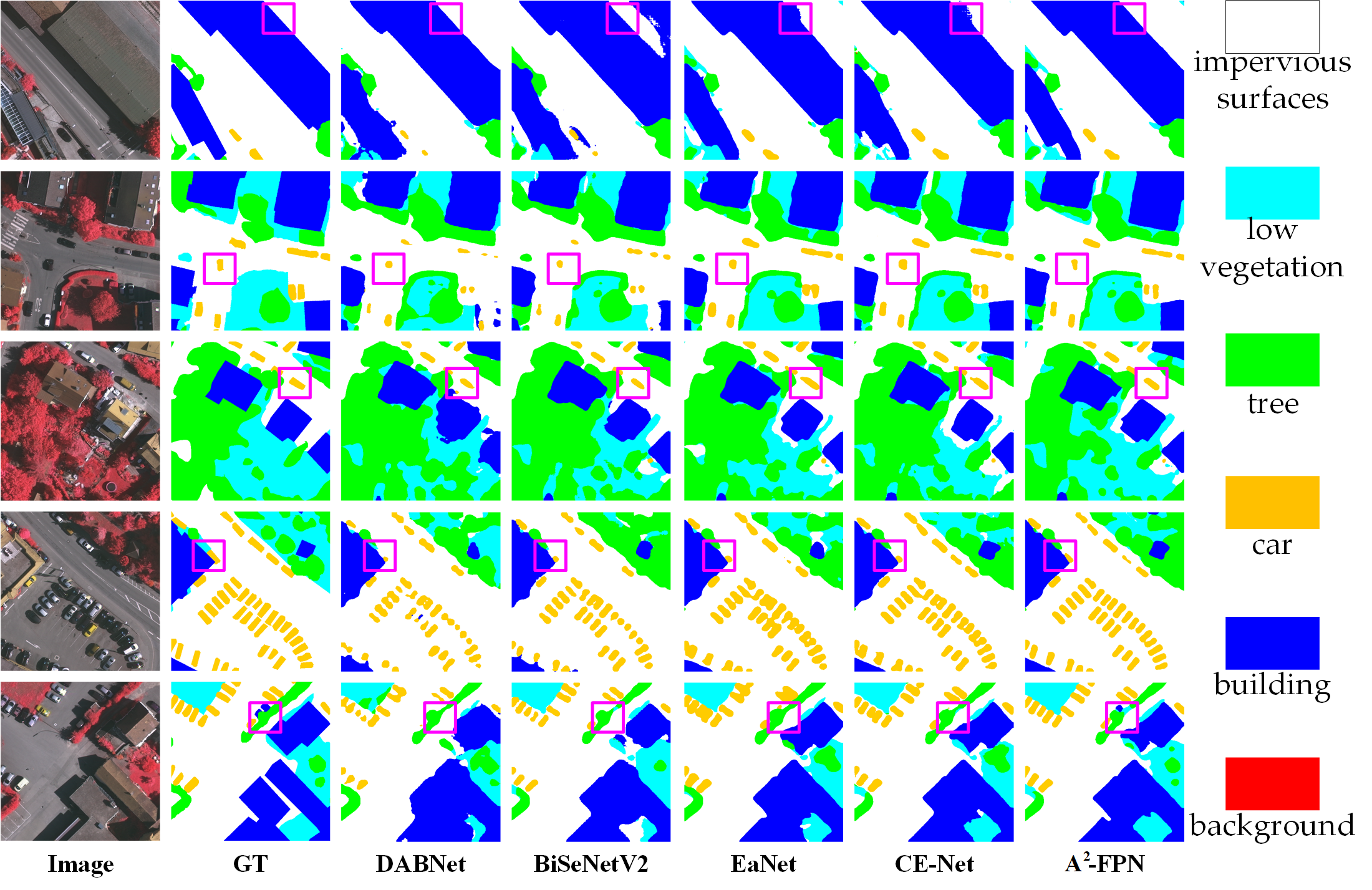}
\caption{Visualization of results on the Vaihingen dataset.}
\label{fig:6}
\end{figure*}

 To qualitatively illustrate the effectiveness of the proposed $ A^2 $-FPN, we provide qualitative comparisons between different networks via $ 512 \times 512 $ patches in Fig. \ref{fig:6}. Particularly, we leverage the red box to mark those intricate regions which are easy to be confused. It can be seen that the elaborate feature pyramid and attention aggregation enable our $ A^2 $-FPN to generate more accurate segmentation maps.

\begin{table*}[]
\setlength{\abovecaptionskip}{0.cm}
\centering
\caption{The Experimental Results on the Vaihingen Dataset.}
\label{table:1}
\begin{tabular}{c|c|ccccc|ccc}
\hline
Method    & Backbone & Imp. surf.      & Building        & Low veg.        & Tree            & Car             & Mean F1         & OA (\%)         & mIoU (\%)       \\ \hline
U-Net     & -        & 84.3          & 86.5          & 73.1          & 83.9          & 40.8          & 73.7          & 82.0          & 64.0          \\
DABNet    & -        & 87.8          & 88.8          & 74.3          & 84.9          & 60.2          & 79.2          & 84.3          & 70.2          \\
BiSeNetV2 & -        & 89.9          & 91.9          & 82.0          & 88.3          & 71.4          & 84.7          & 88.0          & 75.5          \\
PSPNet    & ResNet-34 & 90.3          & 94.2          & 82.8          & 88.6          & 51.1          & 81.4          & 88.8           & 71.3          \\
DANet     & ResNet-34 & 91.1          & 94.8          & 83.5          & 88.9          & 63.0          & 84.3          & 89.5          & 74.4          \\
EaNet     & ResNet-34 & 92.8          & 95.2          & 82.8          & 89.3          & 80.6          & 88.0          & 90.0          & 79.1          \\
CE-Net    & ResNet-34 & 92.7          & 95.5          & 83.4          & 89.5          & 81.2          & 88.5          & 90.4          & 79.7          \\ \hline
$ A^2 $-FPN   & ResNet34 & \textbf{93.0} & \textbf{95.7} & \textbf{84.7} & \textbf{90.0} & \textbf{86.9} & \textbf{90.1} & \textbf{91.0} & \textbf{82.2} \\ \hline
\end{tabular}
\end{table*}

\subsection{Results on the ISPRS Potsdam Dataset}
\par
\begin{figure*}[ht]
\centering
\includegraphics[scale=1.3]{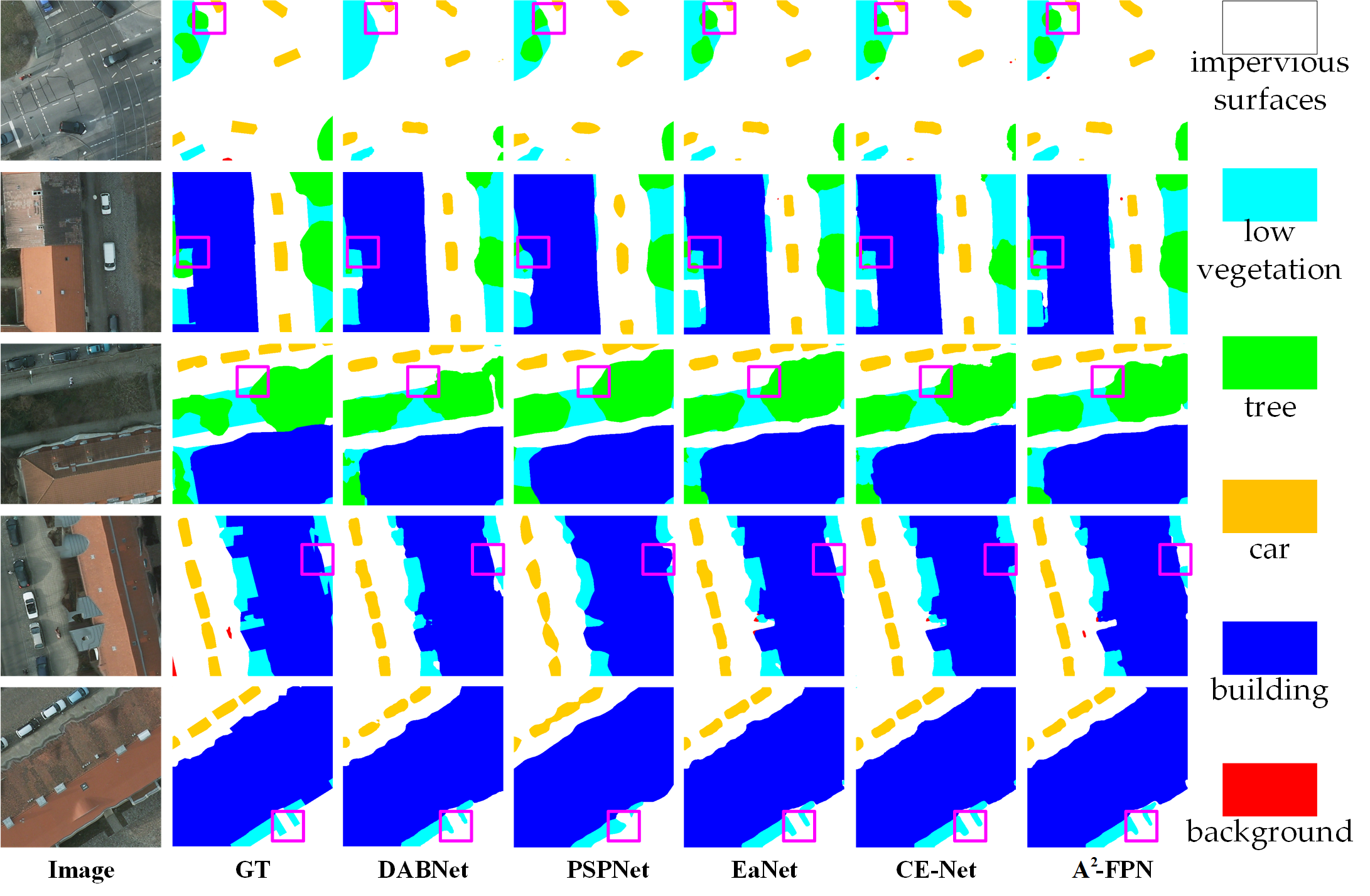}
\caption{Visualization of results on the Potsdam dataset.}
\label{fig:7}
\end{figure*}

 To further evaluate the effectiveness of $ A^2 $-FPN, we carry out experiments on the ISPRS Potsdam dataset. The training and testing settings on the Potsdam dataset are the same as the Vaihingen dataset. Numerical comparisons with comparative algorithms are listed in Table \ref{table:2}. The $ A^2 $-FPN achieves up to 92.4\% in mean F1 score, 91.1\% in overall accuracy, and 86.1\% in mIoU. In Fig. \ref{fig:7}, we further visualize $ 512 \times 512 $ patches with the intractable regions marked by red rectangles. Our $ A^2 $-FPN produces consistently better segmentation results than other benchmark approaches.

\begin{table*}[]
\setlength{\abovecaptionskip}{0.cm}
\centering
\caption{The Experimental Results on the Potsdam Dataset.}
\label{table:2}
\begin{tabular}{c|c|ccccc|ccc}
\hline
Method    & Backbone & Imp. surf.      & Building        & Low veg.        & Tree            & Car             & Mean F1         & OA (\%)         & mIoU (\%)       \\ \hline
U-Net     & -        & 85.0          & 88.8          & 76.7          & 73.1          & 90.3          & 82.8          & 80.6          & 74.3          \\
DABNet    & -        & 89.9          & 93.2          & 83.6          & 82.3          & 92.6          & 88.3          & 86.7          & 79.6          \\
BiSeNetV2 & -        & 91.3          & 94.3          & 85.0          & 85.2          & 94.1          & 90.0 & 88.2          & 82.3          \\
PSPNet    & ResNet-34 & 91.6          & 95.8          & 86.0          & 87.7          & 86.5          & 89.5          & 89.5          & 82.6          \\
DANet     & ResNet-34 & 91.9          & 96.1          & 85.6          & 87.6          & 86.8          & 89.6          & 89.6          & 82.6          \\
EaNet     & ResNet-34 & 92.4          & 96.3          & 85.6          & 87.9          & 95.1          & 91.5          & 89.7          & 85.2          \\
CE-Net    & ResNet-34 & 92.5          & 96.4          & 86.4          & 87.8          & 95.3          & 91.7          & 90.0          & 85.4          \\ \hline
$ A^2 $-FPN   & ResNet-34 & \textbf{93.6} & \textbf{96.9} & \textbf{87.5} & \textbf{88.4} & \textbf{95.7} & \textbf{92.4} & \textbf{91.1} & \textbf{86.1} \\ \hline
\end{tabular}
\end{table*}

\subsection{Results on the GID Dataset}
\par
\begin{figure*}[t]
\centering
\includegraphics[scale=1.3]{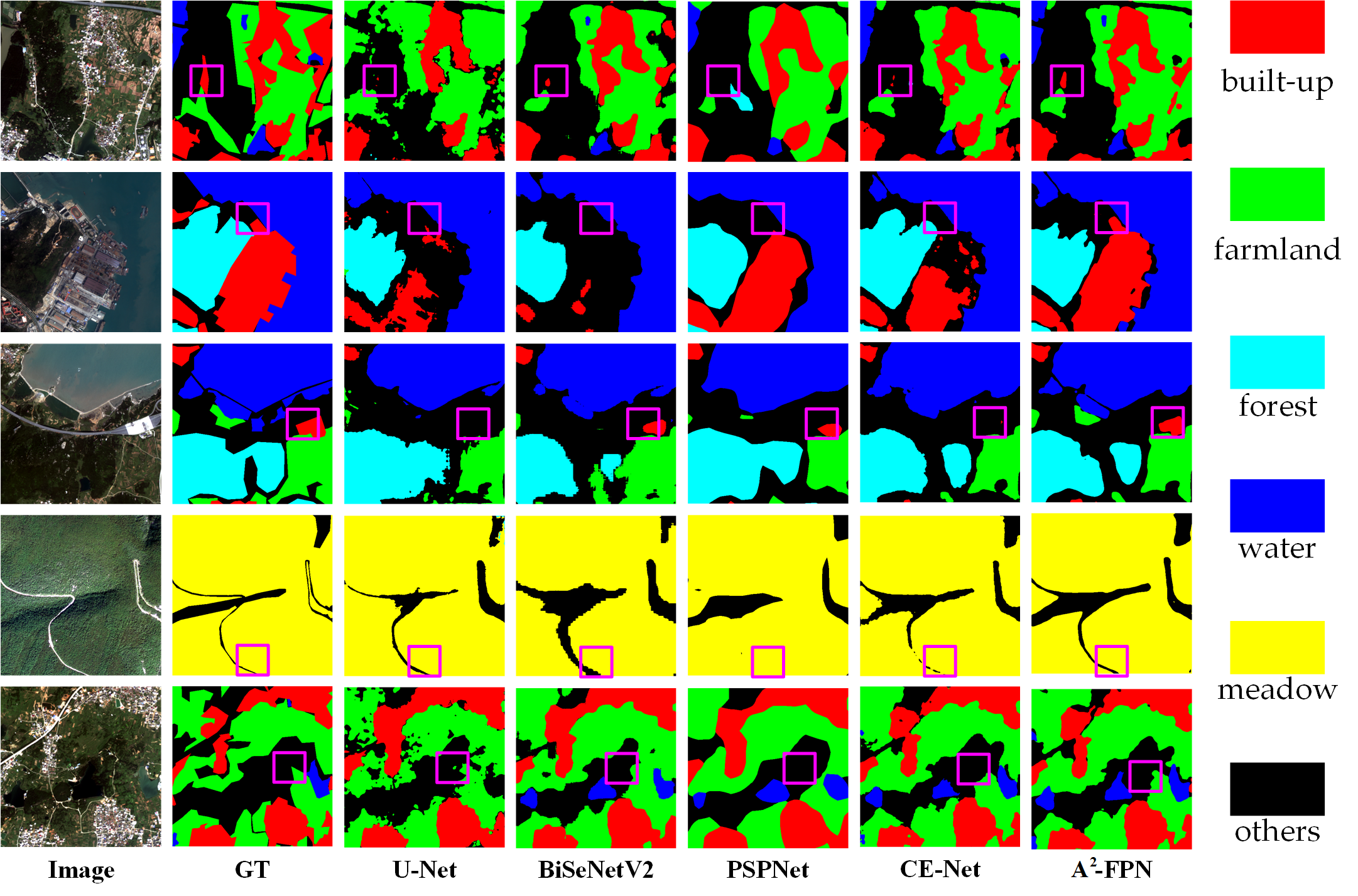}
\caption{Visualization of results on the GID dataset.}
\label{fig:8}
\end{figure*}

 We conducted experiments on the GID dataset to further test the accuracy of our $ A^2 $-FPN. As listed in Table \ref{table:3}, our $ A^2 $-FPN holds the leading position on the vast majority of the evaluation indexes. Visualized results in Fig. \ref{fig:8} also demonstrates the superiority of our method.

\begin{table*}[]
\setlength{\abovecaptionskip}{0.cm}
\centering
\caption{The Experimental Results on the GID Dataset.}
\label{table:3}
\begin{tabular}{c|c|cccccc|ccc}
\hline
Method    & Backbone & build-up        & forest          & farmland        & meadow          & water           & others          & Mean F1         & OA (\%)         & mIoU (\%)       \\ \hline
U-Net     & -        & 82.3          & 85.0          & 89.7          & 84.1          & 93.2          & 69.2          & 83.9          & 82.3          & 73.0          \\
DABNet    & -        & 81.7          & 86.9          & 90.6          & 85.9          & 94.2          & 72.7          & 85.3          & 83.9          & 75.0          \\
BiSeNetV2 & -        & 83.0          & 86.4          & 90.2          & 86.4          & 94.7          & 72.4          & 85.5          & 83.9          & 75.4          \\
PSPNet    & ResNet-34 & 84.2          & 89.1          & 91.5          & 87.6          & 95.1          & 76.4          & 87.3          & 86.1          & 77.9          \\
DANet     & ResNet-34 & 84.8          & 89.5          & 91.7          & 87.8          & 95.6          & 77.8          & 87.9          & 86.7          & 78.8          \\
EaNet     & ResNet-34 & 85.2          & 90.4          & 91.8          & 86.4          & 96.2          & 78.4          & 88.1          & 87.3          & 79.1          \\
CE-Net    & ResNet-34 & 85.9          & 90.2          & 92.2 & 87.4          & 96.5          & 79.4          & 88.6          & 87.7          & 79.9          \\ \hline
$ A^2 $-FPN   & ResNet-34 & \textbf{86.3} & \textbf{91.0} & \textbf{92.4}          & \textbf{87.9} & \textbf{96.8} & \textbf{79.9} & \textbf{89.1} & \textbf{88.3} & \textbf{80.7} \\ \hline
\end{tabular}
\end{table*}

\section{Discussion}
\label{sec:dis}

\subsection{Ablation Study about FPN and AAM}
\begin{figure*}[t]
\centering
\includegraphics[scale=1]{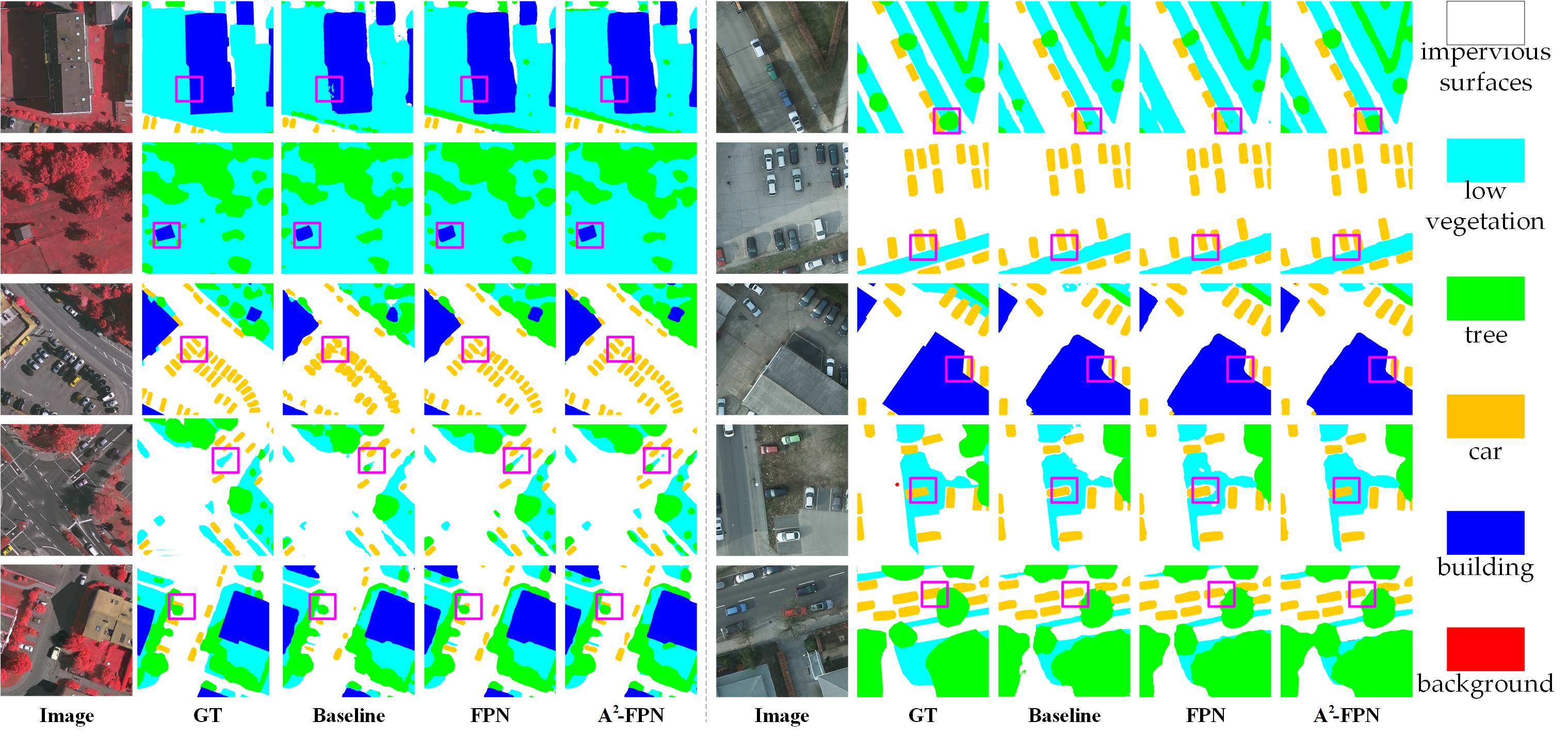}
\caption{Visualization of ablation study on (left) the Vaihingen dataset and (right) the Potsdam dataset.}
\label{fig:9}
\end{figure*}

\begin{figure}[t]
\centering
\includegraphics[scale=1.6]{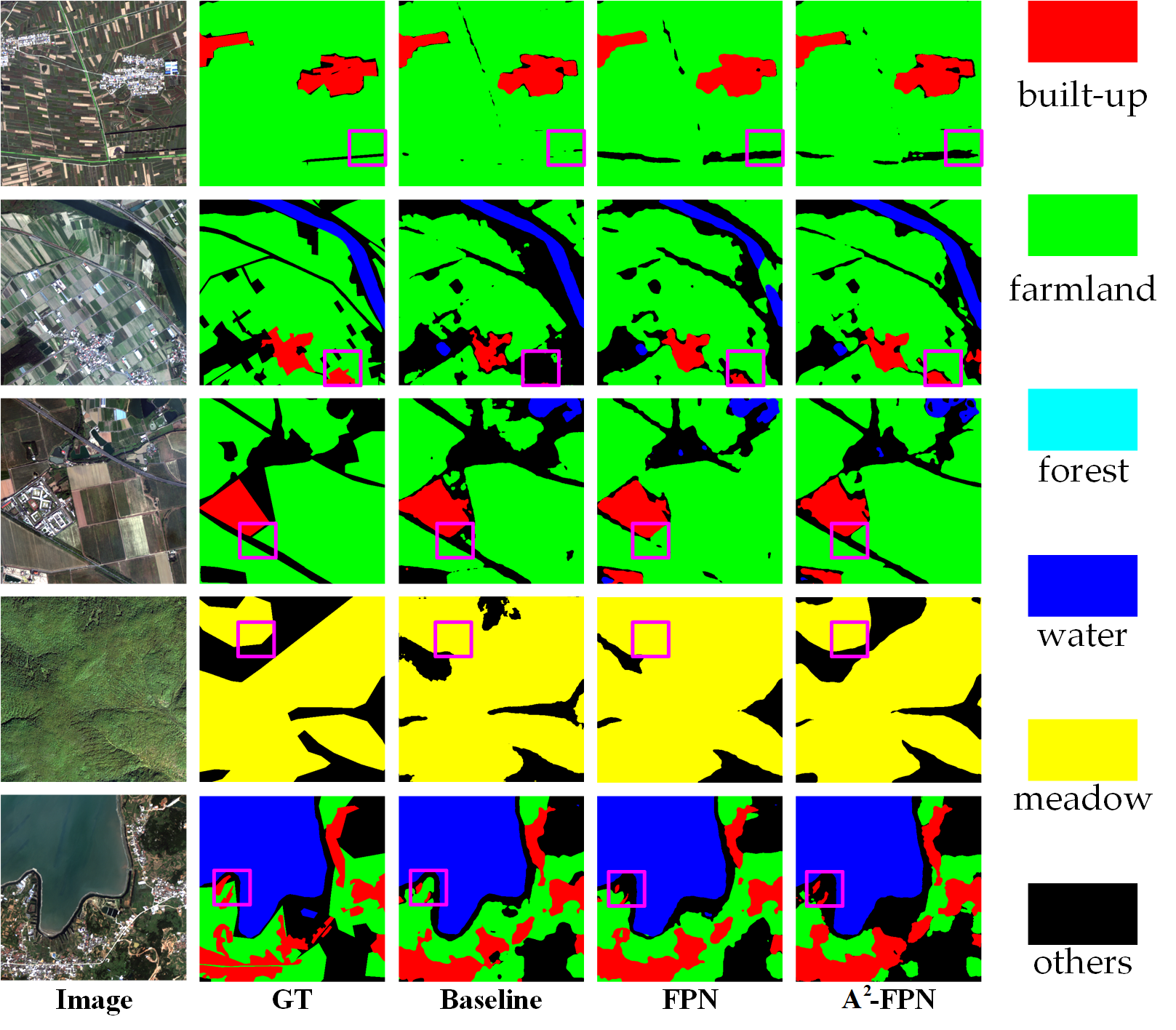}
\caption{Visualization of ablation study on the GID dataset.}
\label{fig:10}
\end{figure}

Ablation experiments were conducted to test the effectiveness of FPN and AAM in the proposed $ A^2 $-FPN. The encoder-decoder structure based on ResNet-34 is selected as the baseline. As shown in Table \ref{table:4}, the FPN outperforms the encoder-decoder baseline significantly. For the Vaihingen dataset, the introduction of FPN brings more than 3.6\% in mean F1 score, 1.1\% in OA, and 3.8\% in mIoU, while the improvements for the Potsdam dataset is 0.6\%, 0.7\%, and 2.7\%, respectively. The FPN is initially designed for object detection. To tackle the segmentation issue, the feature maps generated by feature pyramids are simply concatenated, lacking the global context information crucial for segmentation. Therefore, the Attention Aggregation Module is developed to address the above limitation. As a specifically designed module for semantic segmentation, the utilization of AAM contributes to the increase of more than 0.6\% in mean F1 score, 0.6\% in OA, and 0.9\% in mIoU for the Vaihingen dataset, while the figures for the Potsdam dataset are about 0.7\%, 0.9\%, and 0.7\%, respectively. For qualitative comparison, we visualize certain segmentation maps generated by the baseline, FPN, and our $ A^2 $-FPN, which can be seen from Fig. \ref{fig:9}. Besides, the increases brought by the AAM on the GID dataset are about 0.7\% in mean F1 score, 0.8\% in OA, and 1.0\% in mIoU, and the visualization results are shown in Fig. \ref{fig:10}.

\begin{table}[]
\setlength{\abovecaptionskip}{0.cm}
\centering
\caption{Ablation Study about FPN and AAM.}
\label{table:4}
\begin{tabular}{c|cc|ccc}
\hline
Dataset                    & Method   & Backbone & Mean F1 & OA     & mIoU   \\ \hline
\multirow{3}{*}{Vaihingen} & Baseline & ResNet-34 & 85.9  & 89.5 & 77.5 \\
                           & FPN      & ResNet-34 & 89.5  & 90.4 & 81.3 \\
                           & $ A^2 $-FPN  & ResNet-34 & 90.1  & 91.0 & 82.2 \\ \hline
\multirow{3}{*}{Potsdam}   & Baseline & ResNet-34 & 91.1  & 89.5 & 82.7 \\
                           & FPN      & ResNet-34 & 91.7  & 90.2 & 85.4 \\
                           & $ A^2 $-FPN  & ResNet-34 & 92.4  & 91.1 & 86.1 \\ \hline
\multirow{3}{*}{GID}       & Baseline & ResNet-34 & 87.4  & 86.1 & 78.0 \\
                           & FPN      & ResNet-34 & 88.4  & 87.5 & 79.7 \\
                           & $ A^2 $-FPN  & ResNet-34 & 89.1  & 88.3 & 80.7 \\ \hline
\end{tabular}
\end{table}

\subsection{Limitation}
\par Although the proposed $ A^2 $-FPN has bridged the gap between low-level and high-level features and compensated for the weakness of the raw FPN, there are still some potential issues that need to be considered. First, the total trainable parameters in the $ A^2 $-FPN are 22.93M, which is less than medium-scale networks such as DANet (24.96M), PSPNet (34.14M), and EaNet (44.34M) while larger than those small-scale networks such as BiSeNetV2 (12.30M). Second, the incorporation of auxiliary information (e.g. DSMs) might further increase the accuracy. However, these require intelligent approaches to handle computationally intensive operations to include more information. Our future work will, therefore, devote to realizing real-time semantic segmentation, as well as developing efficient techniques to fuse DSMs or nDSMs, thereby further enhancing the segmentation performance.

\section{Conclusion}
\par The automatic semantic segmentation from fine-resolution remotely sensed images remains a complicated and challenging task, due to the limited spatial and contextual information utilized. In this research, we employ the Feature Pyramid Network to combine the extracted spatial and contextual features comprehensively. In particular, the pyramidal hierarchy enables FPN to combine low-level detailed spatial information with high-level abundant semantic features thoroughly. Besides, to enhance the segmentation accuracy, we propose an Attention Aggregation Module to not only merge the feature maps generated by the feature pyramid effectively but also to extract the global context information fully. Substantial experiments conducted on the ISPRS Vaihingen, Potsdam, and GID datasets demonstrate the effectiveness of our $ A^2 $-FPN. The extensive ablation studies illustrate the validity of FPN and AAM accordingly.

\vbox{}
\noindent \textbf{Funding} This work was supported in part by the National Natural Science Foundation of China (No. 41671452). 

\vbox{}
\noindent \textbf{Availability of Data and Material} The data used to support the findings of this study are included within the article.

\vbox{}
\noindent \textbf{Code Availability} Software applications or custom code generated or used during the study are available at GitHub. 

\vbox{}
\noindent \textbf{Conflict of Interests} The authors have no conflicts of interest to declare that are relevant to the content of this article.

\vbox{}
\noindent \textbf{Author's Contributions} This work was conducted in collaboration with all author. Shunyi Zheng supervised the research work and provided experimental facilities. Rui Li and Chenxi Duan designed the semantic segmentation model and conducted the experiments. This manuscript was written by Rui Li and Chenxi Duan. Ce Zhang and Libo Wang checked the experimental results. All authors have read and agreed to the published version of the manuscript.

%
%


\bibliographystyle{IEEEtran}
\bibliography{IEEEabrv,A2-FPN}

\end{document}